\documentclass[10pt,twocolumn,letterpaper]{article}

\usepackage[pagenumbers]{cvpr} 

\usepackage{anyfontsize}
\usepackage{bbding}
\usepackage{graphicx}
\usepackage{makecell}

\usepackage{colortbl}
\usepackage{multirow}

\usepackage[ruled,vlined]{algorithm2e}
\usepackage{algpseudocode}
\renewcommand{\algorithmicrequire}{\textbf{Input:}} 
\renewcommand{\algorithmicensure}{\textbf{Output:}}
\definecolor{cvprblue}{rgb}{0.21,0.49,0.74}
\usepackage[pagebackref,breaklinks,colorlinks,allcolors=cvprblue]{hyperref}

\title{Dual-view X-ray Detection: Can AI Detect Prohibited Items from Dual-view X-ray Images like Humans?}

\author{
Renshuai Tao$^{1}$, Haoyu Wang$^{1}$, Yuzhe Guo$^{1}$, Hairong Chen$^{1}$, Li Zhang$^{3}$\\ Xianglong Liu$^{2}$, Yunchao Wei$^{1}$, Yao Zhao$^{1}$\\
\textsuperscript{1}{Beijing Jiaotong University}, \textsuperscript{2}{Beihang University}, \textsuperscript{3}{Tsinghua University}\\ {\fontsize{8.5pt}{\baselineskip}\selectfont \tt rstao@bjtu.edu.cn}
}

\begin{document}
\maketitle

\begin{abstract}
To detect prohibited items in challenging categories, human inspectors typically rely on images from two distinct views (vertical and side). Can AI detect prohibited items from dual-view X-ray images in the same way humans do? Existing X-ray datasets often suffer from limitations, such as single-view imaging or insufficient sample diversity. To address these gaps, we introduce the Large-scale Dual-view X-ray (LDXray), which consists of 353,646 instances across 12 categories, providing a diverse and comprehensive resource for training and evaluating models. To emulate human intelligence in dual-view detection, we propose the Auxiliary-view Enhanced Network (AENet), a novel detection framework that leverages both the main and auxiliary views of the same object. The main-view pipeline focuses on detecting common categories, while the auxiliary-view pipeline handles more challenging categories using ``expert models" learned from the main view. Extensive experiments on the LDXray dataset demonstrate that the dual-view mechanism significantly enhances detection performance, e.g., achieving improvements of up to \textbf{+24.7\%} for the challenging category of umbrellas. Furthermore, our results show that AENet exhibits strong generalization across seven different detection models for X-ray Inspection\footnote{The code and dataset will be released upon acceptance.}.

\end{abstract}
\vspace{-0.1in}
\section{Introduction}
Security inspectors typically use X-ray scanners to detect prohibited items hidden in luggage, a process that is both time-consuming and labor-intensive. The development of deep learning technologies\cite{zeng2024collapsed,mallik2024priorband,guo2023cbanet,guo2023multidimensional}, particularly in computer vision\cite{zhao2023cddfuse,xu2021deep,zhao2022discrete}, is expected to reduce the workload through automatic prohibited items detection.

\begin{figure}[!t]
  \centering
\includegraphics[width=\linewidth]{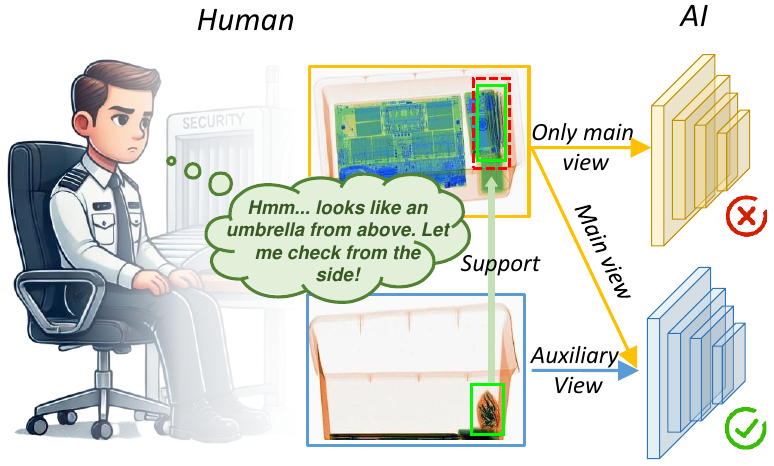}
\vspace{-0.2in}
   \caption{Illustration of the dual-view X-ray detection task. This paper aims to develop AI capable of reasoning like human experts when detecting prohibited items, leveraging insights from dual-view X-ray images to improve accuracy and interpretative ability.}
   \label{fig:onecol}
   \vspace{-0.15in}
\end{figure}
In recent years, detecting prohibited items in X-ray security scenes has become a research hotspot, leading to the development of numerous datasets and detection methods\cite{GDXray,SIXray,OPIXray,Hixray,EDS,DvXray}. Datasets include a considerable number of categories and instances, offering a comprehensive resource for training and evaluating models\cite{kopf2024openassistant,si2024spokenwoz,cobb2023aircraftverse,kaltenborn2023climateset,liu2024largest,shi2024m5hisdoc,he2024species196,gulino2024waymax,li2024scenarionet}. These efforts have attracted the attention of researchers in this field, resulting in significant advancements and increased collaboration.

In real-world scenarios, human inspectors typically rely on images from two distinct views (vertical and side) to effectively detect prohibited items, particularly those in challenging categories. This dual-view approach is crucial because single-view images often fail to capture vital information that can be seen from the other perspective. The limitation of single-view images becomes especially significant when objects are randomly placed in luggage, leading to heavy occlusions and making it difficult to identify hidden items. In such cases, relying solely on a single view can result in missed detections. The question then arises: \textbf{Can AI detect prohibited items from dual-view X-ray images in the same way humans do, leveraging both views for enhanced detection?} Achieving this goal requires not only a large-scale evaluation dataset but also the development and deployment of sophisticated, professional models that can effectively process and interpret dual-view data.

Despite these challenges, as illustrated in Table \ref{dataset-compare}, existing X-ray datasets often suffer from the limitations of single-view imaging or insufficient sample diversity. To address these gaps, in this paper, we introduce a Large-scale Dual-view X-ray (LDXray) dataset for prohibite items detection task, sourced directly from real-world security environments.
LDXray contains 146,997 paired images, making it approximately 20 times larger than similar existing X-ray datasets, and represents 353,646 instances across 12 categories.
We employed professional human security inspectors to ensure high annotation accuracy.
We anonymized all personal information, such as time and location, to prevent linkage of the luggage to owners, guaranteeing data privacy.

To further improve detection performance, we propose a novel framework called the Auxiliary-view Enhanced Network (AENet). AENet is designed to leverage both main and auxiliary views of the same object, improving object detection in X-ray security inspection. The AENet framework consists of two distinct detection pipelines: (1) The main view pipeline is responsible for detecting simple categories, using a universal model applicable to all standard categories. (2) The auxiliary view pipeline utilizes saliency detection techniques to identify challenging objects and then hands over these candidates to specialized ``expert models" for further analysis and confirmation. The outputs from both pipelines are combined to produce the final detection results. \textbf{The proposed  dual-pipeline approach enhances the detection of challenging categories while maintaining stable performance for other simpler categories}.

We conduct extensive experiments on the proposed LDXray dataset and compare the performance of various widely-used detection models. Our results demonstrate that the dual-view mechanism significantly enhances detection accuracy, especially for challenging categories, \emph{e.g.}, improving from 33.9\% to 58.6\% (\textbf{+24.7\%}) for the umbrella category. Additionally, AENet shows strong generalization across different base models, with seven detection models implemented. We hope that future research will further explore the unique characteristics of dual-view data, leading to the development of more intelligent security inspection solutions. In summary, the key contributions are as follows:

\begin{itemize}
\item We introduce the Large-scale Dual-view X-ray (LDXray) dataset, a significant contribution to the field of X-ray security inspection. LDXray consists of a diverse and comprehensive set of dual-view X-ray images, specifically designed for the detection of prohibited items in luggage.
\item We propose a novel detection framework, AENet, that leverages both the main and auxiliary views of the same object to significantly enhance detection, especially for challenging categories. The core idea is to combine the advantages of both main and auxiliary view.
\item We conduct extensive experiments on the newly introduced LDXray dataset and compare our method with popular methods. Experimental results clearly show that this dual-view mechanism provides a significant performance improvement in detecting the challenging categories.
\end{itemize}

The remainder of the paper is organized as follows: Section \ref{Section:relatedwork} reviews related work on X-ray prohibited items detection task. Section \ref{Section:dataset} provides an overview of the proposed LDXray dataset. Section \ref{Section:method} details the proposed AENet method. In Section \ref{sec:exp}, we present the experimental setup and results. Finally, Section \ref{Section:conclusion} offers concluding remarks.

\begin{table}[!b]
\vspace{-0.1in}
\setlength{\tabcolsep}{0.36mm}
  \fontsize{8.1}{13.8}\selectfont
  \centering
  \begin{tabular}{lcccccccc}
    \toprule
    Dataset & 2-view & Open & $N_{c}$ & $N_{p}$ & $N_{a}$ & Task & Year  \\
    \midrule
    GDXray~\cite{GDXray} & \textcolor[rgb]{1,0,0}{\XSolidBrush} &  \textcolor[rgb]{0,0.5,0}{\Checkmark} & 3 & 8,150 & 8,150 & Detection & 2015 \\ 
    SIXray~\cite{SIXray}  & \textcolor[rgb]{1,0,0}{\XSolidBrush} & \textcolor[rgb]{0,0.5,0}{\Checkmark}& 6 & 8,929 & 8,929 & Cls \& Det & 2019  \\
    OPIXray~\cite{OPIXray} &  \textcolor[rgb]{1,0,0}{\XSolidBrush} & \textcolor[rgb]{0,0.5,0}{\Checkmark} & 5 & 8,885 & 8,885 & Detection  & 2020  \\
    PIDray~\cite{PIDray} &  \textcolor[rgb]{1,0,0}{\XSolidBrush} &\textcolor[rgb]{0,0.5,0}{\Checkmark} & 12 & 47,677 & 47,677 & Det \& Seg & 2021  \\
    HiXray~\cite{Hixray} & \textcolor[rgb]{1,0,0}{\XSolidBrush} & \textcolor[rgb]{0,0.5,0}{\Checkmark} & 8 & 45,364 & 102,928 & Detection & 2021 \\
    PIXray~\cite{PIXray} & \textcolor[rgb]{1,0,0}{\XSolidBrush} & \textcolor[rgb]{0,0.5,0}{\Checkmark} & 15 & 5,046 & 15,201 & Det \& Seg & 2022 \\ 
    CLCXray~\cite{CLCXray} & \textcolor[rgb]{1,0,0}{\XSolidBrush} & \textcolor[rgb]{0,0.5,0}{\Checkmark} & 12 & 9,565 & 9,565 & Detection &  2022 \\
    EDS~\cite{EDS} & \textcolor[rgb]{1,0,0}{\XSolidBrush} & \textcolor[rgb]{0,0.5,0}{\Checkmark} & 10 & 14,219 & 31,654 & Detection & 2022 \\
    FSOD~\cite{FSOD} & \textcolor[rgb]{1,0,0}{\XSolidBrush} & \textcolor[rgb]{0,0.5,0}{\Checkmark} & 20 & 12,333 & 41,704 & Detection & 2022 \\
    LPIXray~\cite{LPIXray} & \textcolor[rgb]{1,0,0}{\XSolidBrush}  & \textcolor[rgb]{1,0,0}{\XSolidBrush} & 18 & 60,950 & 84,785 & Detection & 2023 \\
    \midrule
    MV-Xray~\cite{mvxray} & \textcolor[rgb]{0,0.5,0}{\Checkmark} & \textcolor[rgb]{1,0,0}{\XSolidBrush} & 2 & 3,231 & 3,231 & Detection & 2019 \\
    deei6~\cite{deei6} & \textcolor[rgb]{0,0.5,0}{\Checkmark} & \textcolor[rgb]{1,0,0}{\XSolidBrush} & 6 & 7,022 & 7,022 & Det \& Seg &2021 \\
    DBF~\cite{DBF} & \textcolor[rgb]{0,0.5,0}{\Checkmark} & \textcolor[rgb]{1,0,0}{\XSolidBrush} & 4 & 2,528 & 3,034 & Detection & 2021  \\
    Dualray~\cite{dualray} & \textcolor[rgb]{0,0.5,0}{\Checkmark} & \textcolor[rgb]{1,0,0}{\XSolidBrush} & 6 & 4,371 & 4,371 & Detection & 2022 \\
    DvXray~\cite{DvXray} & \textcolor[rgb]{0,0.5,0}{\Checkmark} & \textcolor[rgb]{0,0.5,0}{\Checkmark} & 15 & 5,000 & 5,496 & Classification & 2024  \\
    \midrule
    LDXray (\textbf{ours}) & \textcolor[rgb]{0,0.5,0}{\Checkmark} & \textcolor[rgb]{0,0.5,0}{\Checkmark} & 12 & \textbf{146,997} & \textbf{353,646} & Detection & 2024  \\
    \bottomrule
  \end{tabular}
    \caption{
  A comprehensive overview of various X-ray datasets utilized in security inspection. The number of categories ($N_{c}$), images containing prohibited items ($N_{p}$), and annotations ($N_{a}$) associated are displayed. Additionally, we summary the task, year, and availability of each X-ray dataset for a more complete comparison.
  }
\label{dataset-compare}
\end{table}

\begin{figure*}[!t]
\centering
\vspace{-0.35in}
\includegraphics[width=\linewidth]{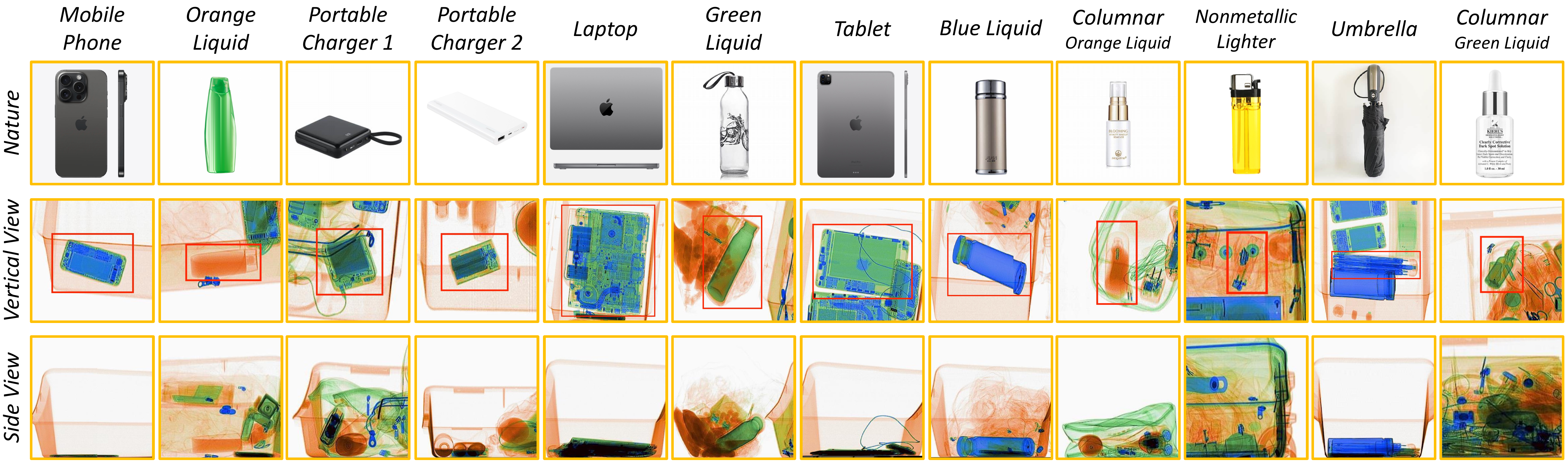}
\vspace{-0.25in}
\caption{Various categories of objects and corresponding X-ray images from two different views in LDXray. The top row displays natural images of the objects. The middle row shows the X-ray images taken from the vertical view, while the bottom row shows the side view.}
\vspace{-0.2in}
\label{fig:dataset}
\end{figure*}
\section{Related Work}\label{Section:relatedwork}

The development of X-ray image analysis for security and inspection has been greatly supported by a range of specialized datasets. Foundational datasets like GDXray \cite{GDXray} and SIXray \cite{SIXray} initially established resources for detection, while more recent datasets such as PIDray \cite{PIDray}, HiXray \cite{Hixray}, and PIXray \cite{PIXray} have expanded task diversity with comprehensive annotations. Dual-view datasets, including MV-Xray \cite{mvxray} and Dualray \cite{dualray}, introduced multi-view data, enhancing depth in scene analysis. Our proposed dataset, LDXray, offers the largest annotated image set to date, supporting both dual-view and open-access needs for data diversity and task applicability. \textbf{Detailed analysis are provided in the Supplementary Materials}.

\section{LDXray Dataset}
\label{Section:dataset}

We introduce LDXray, the first large-scale dual-view X-ray dataset designed for prohibited items detection in X-ray images. In this section, we present a comprehensive overview of the construction details and statistics of LDXray dataset.

\subsection{Construction Details}\label{constraction-details}

\textbf{Data Collection.} Due to significant differences between the distribution of artificial and real-world scenarios, such as the appearance frequency of certain categories and their co-occurrence with specific items, we collected the images from actual transportation hubs (to protect privacy, hubs are anonymous). When packages undergo inspections, the X-ray machine generates and stores corresponding X-ray images. We collect a large number of images, subsequently removing low-quality and invalid samples. The long-tail distribution accurately matched the real-world scenario.

\textbf{Privacy Protect.} We conduct a two-step process to safeguard privacy. First, during the period of collection, an obvious indicator was hung to inform the passengers that the dataset may be used for public academical research after personal information removed. \textbf{And security inspectors also asked passengers for their explicit consent.} If consent was not given, the X-ray image was deleted. Additionally, we have de-identified the collected data by removing details such as time and location that could potentially identify an individual. This ensures that the data does not contain Personally Identifiable Information (PII) \cite{mccallister2010identifiable}.

\textbf{Extensive Diversity.} As illustrated in Figure \ref{fig:dataset}, LDXray contains a wide range of prohibited items, categorized into 12 different classes, including ``Mobile Phone'', ``Orange Liquid", ``Portable Charger 1 (lithium-ion prismatic cell)'', ``Portable Charger 2 (lithium-ion cylindrical cell)'', ``Laptop'', ``Green Liquid'', ``Tablet'', ``Blue Liquid'', ``Columnar Orange Liquid'', ``Nonmetallic Lighter'', ``Umbrella'' and     ``Columnar Green liquid'' (``MP'', ``OL'', ``PC1'', ``PC2'', ``LA'', ``GL'', ``TA'', ``BL'', ``CO'', ``NL'', ``UM'' and ``CG'', for short). This diversity mirrors the variety of items that security inspectors encounter during inspections. We offer a comprehensive dataset that facilitates the development of detection models, to manage diverse scenarios effectively.

\textbf{Annotation by Professional Human X-ray Security Inspectors.} Recognizing objects in X-ray images necessitates specialized training and expertise. The LDXray dataset is manually annotated at the bounding-box level by professional human X-ray security inspectors with extensive daily experience. To guarantee high-quality annotation, we followed the similar quality control procedure of annotation as the famous Pascal VOC \cite{everingham2010pascal}. Specifically, all X-ray security inspectors received detailed annotation guidelines that explained what to annotate, how to draw bounding boxes, and how to handle occlusions. Subsequently, another inspector verified the accuracy of each annotation, checking for omitted objects and ensuring comprehensive labeling.

\subsection{Data Statistics}\label{data-statistics}

\begin{figure}[!t]
    \centering
        \includegraphics[width=\linewidth]{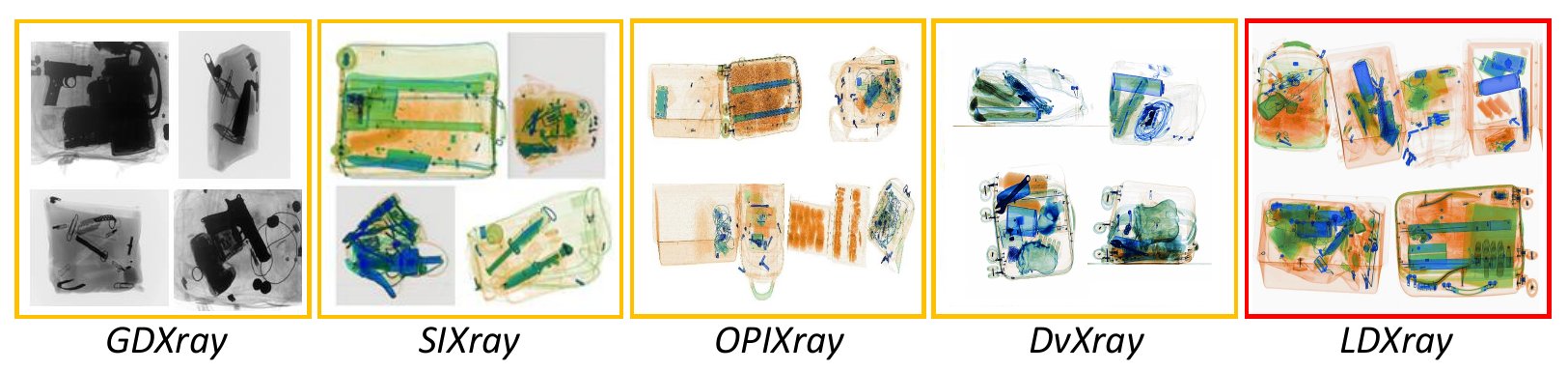}
        \vspace{-0.25in}
        \caption{Image quality comparison of current X-ray datasets. The images of LDXray are superior, attributed to more advanced imaging techniques, enhanced color vibrancy, and higher resolution.}
        \label{fig:dataset_comp}
        \vspace{-0.2in}
\end{figure}

\textbf{Image Quality.} All images of LDXray are stored with an average resolution of 1200x900 pixels, while the maximum resolution of some samples reaches 2000x1040 pixels. As illustrated in Figure \ref{fig:dataset_comp}, the image quality of LDXray is superior, attributed to more advanced imaging techniques and higher resolution. These results demonstrate that LDXray dataset significantly outperforms those popular datasets, enabling a more realistic evaluation for X-ray detection.

\begin{table}[!h]
\setlength{\tabcolsep}{0.46mm}
  \label{dataset info}
  \fontsize{5.5}{11.8}\selectfont
  \centering
  \begin{tabular}{lccccccccccccc}
    \toprule
    Category & MP & OL & PC1 & PC2 & LA & GL & TA & BL & CO & NL & UM & CG & Total \\
    \midrule
    Training & 115,016  & 33,691  & 30,222  & 18,471  & 28,832  & 20,199  & 12,094  & 3,138  & 1,357  & 981  & 683  & 647 & 265,331 \\
    Testing & 37,993  & 11,357  & 10,049  & 6,200  & 9,484  & 6,856  & 4,083  & 1,079  & 445  & 347  & 196  & 226 & 88,315 \\
    \midrule
    Total & 153,009  & 45,048  & 40,271  & 24,671  & 38,316  & 27,055  & 16,177  & 4,217  & 1,802  & 1,328  & 879  & 873& 353,646 \\
    \bottomrule
  \end{tabular}
  \caption{The number of instances in the training and testing sets.}
  \label{dataset info}
  \vspace{-0.1in}
\end{table}

\begin{figure}[!t]
\vspace{-0.05in}
    \centering
    \begin{minipage}{0.115\textwidth}
        \centering
        \begin{subfigure}{\textwidth}
            \centering
            \includegraphics[width=\textwidth]{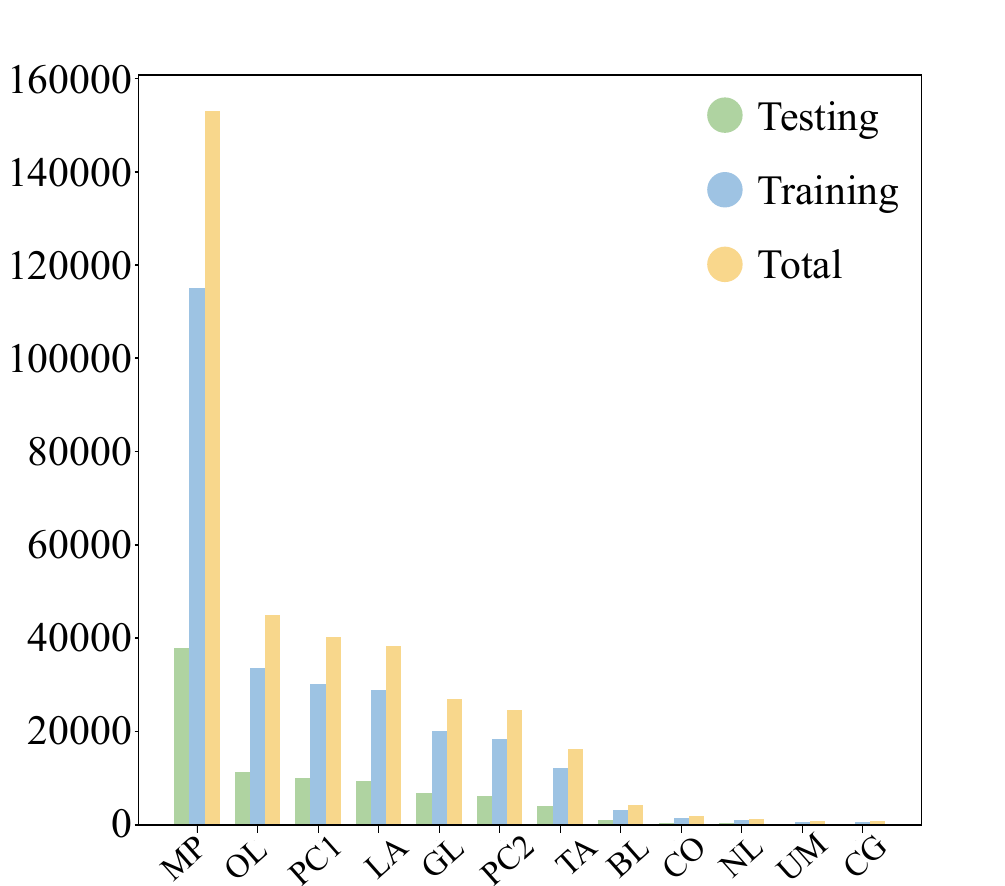}
            \caption{}
            \label{fig:sub1}
        \end{subfigure}
    \end{minipage}
    \hfill
    \begin{minipage}{0.115\textwidth}
        \centering
        \begin{subfigure}{\textwidth}
            \centering
            \includegraphics[width=\textwidth]{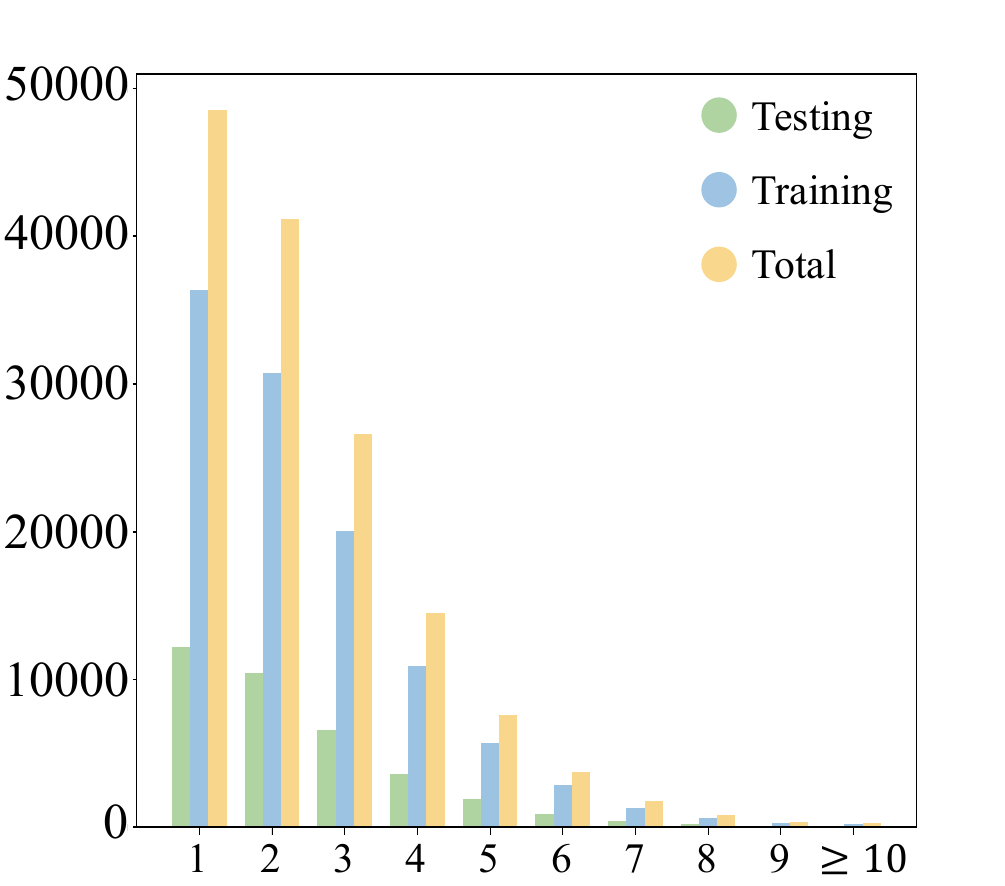}
            \caption{}
            \label{fig:sub2}
        \end{subfigure}
    \end{minipage}
    \hfill
    \begin{minipage}{0.115\textwidth}
        \centering
        \begin{subfigure}{\textwidth}
            \centering
            \includegraphics[width=\textwidth]{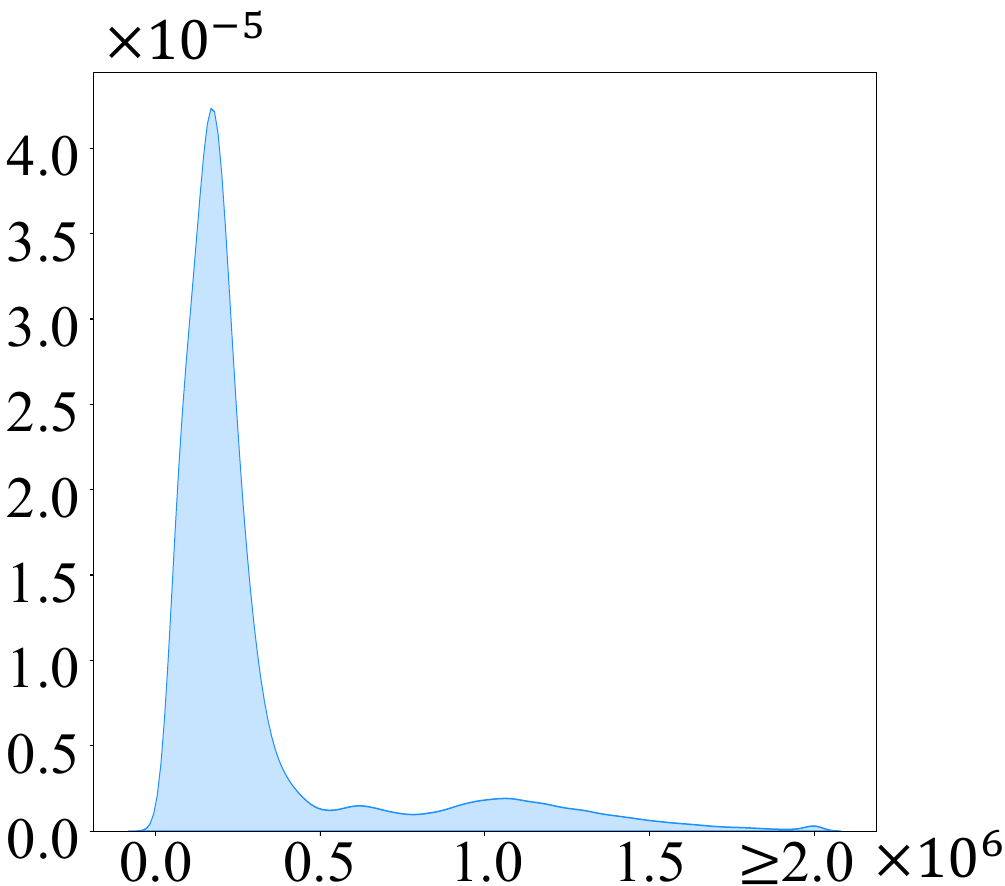}
            \caption{}
            \label{fig:sub3}
        \end{subfigure}
    \end{minipage}
    \hfill
    \begin{minipage}{0.115\textwidth}
        \centering
        \begin{subfigure}{\textwidth}
            \centering
            \includegraphics[width=\textwidth]{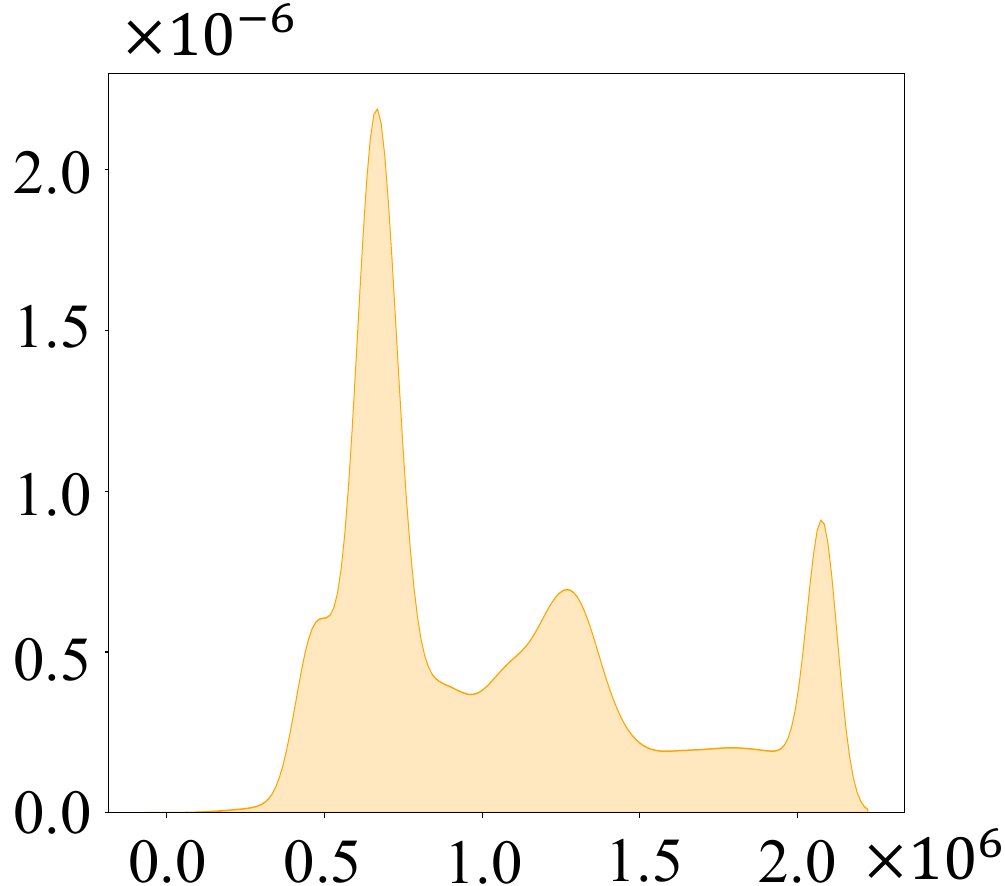}
            \caption{}
            \label{fig:sub4}
        \end{subfigure}
    \end{minipage}
    \vspace{-0.15in}
    \caption{Various distributions. (a) Instance per Category Distribution. (b) Instance per Image Distribution. (c) Instance Area Density Distribution. (d) Image Area Density Distribution.}
    \vspace{-0.15in}
    \label{figure-4-sub}
\end{figure}

\textbf{Category Distribution.} 
LDXray contains 146,997 X-ray image pairs, with 353,646 instances of 12 common categories. The category names and their abbreviations are depicted in ``Construction Details'' section. The dataset is partitioned into a training set and a testing set, where the ratio is about 3:1. The statistics of category distribution of the training set and testing set are also shown in Table \ref{dataset info} and Figure \ref{fig:sub1}, providing a understanding of scale and diversity.

\textbf{Other Distributions.} Several other distributions need to be introduced to further elucidate the characteristics of the LDXray dataset: (1) \textbf{Instance per Image Distribution.} As illustrated in Figure \ref{fig:sub2}, the average number of instances per image in the LDXray dataset is 2.27, with most images not exceeding seven instances. (2) \textbf{Instance Area Density Distribution.} Figure \ref{fig:sub3} illustrates the distribution of instance areas in terms of pixel count. The horizontal axis represents the area of bounding boxes (in square pixels), while the vertical axis indicates density. Most instances have an area below 500,000 square pixels, with only a small fraction exceeding 1,500,000 square pixels. (3) \textbf{Image Area Density Distribution.} Figure \ref{fig:sub4} shows the distribution of image areas, measured in pixel count. The horizontal axis corresponds to the total pixel area of images, and the vertical axis represents density. The majority of images have areas ranging from 500,000 to 2,000,000 square pixels.

\begin{figure*}[!t]
\vspace{-0.35in}
\centering
\includegraphics[width=\linewidth]{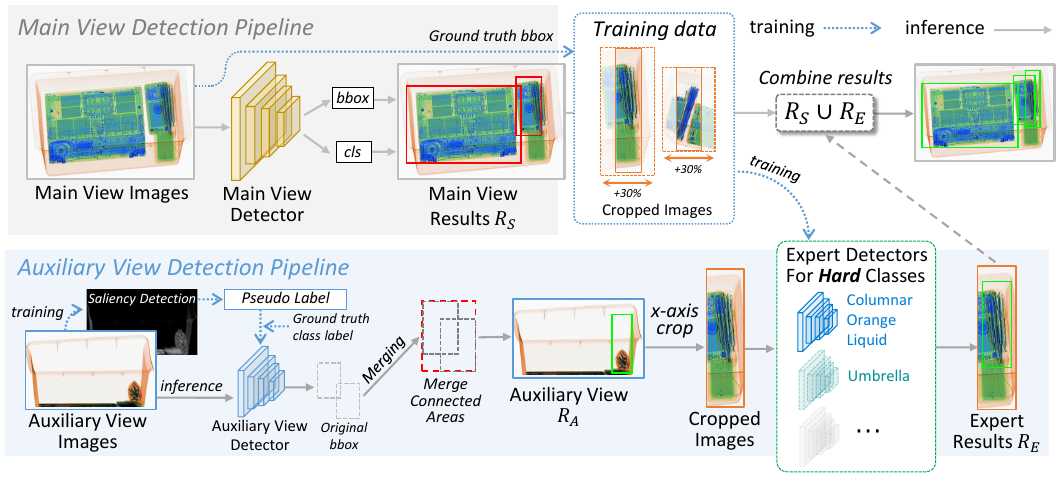}
\vspace{-0.3in}
\caption{Overview of the AENet framework. This architecture consists of two distinct pipelines: the main view, which follows a traditional object detection paradigm, and the auxiliary view, which uses saliency detection and cross-view correspondence for localization.}
\vspace{-0.2in}
\label{fig-method}
\end{figure*}
\section{Method}
\label{Section:method}
In this section, we elaborate on the details of the proposed \textbf{A}uxiliary-view \textbf{E}nhanced \textbf{N}etwork (AENet). We begin with an analysis of the dual-view detection task setting in Section \ref{task_overview}. Next, we introduce the framework of AENet in Section \ref{framework}. We then describe the Main View Detection Pipeline and Auxiliary View Detection Pipeline in Sections \ref{main-view} and \ref{auxiliary-view}, respectively. Finally, in Section \ref{training_progress}, we provide a detailed illustration of the network training process.
\subsection{Task Overview} \label{task_overview}
In the proposed dual-view detection task, the input images consist of a main view and an auxiliary view, denoted as $I_{1}$ and $I_{2}$. The goal is to predict a set of bounding boxes $B = \{b_1, b_2, \dots, b_n\}$ and corresponding class labels $C = \{c_1, c_2, \dots, c_n\}$, where each bounding box $b_i$ defines the spatial extent of an object in the image, and each class label $c_i$ identifies the object category, similar to the traditional object detection task described above. It is important to note that due to the difficulty for inspectors in annotating the auxiliary view, there is no ground truth available for $I_{2}$. Additionally, due to the positional invariance of the imaging devices, a corresponding horizontal coordinate relationship exists between the objects in these two perspective images.

\begin{equation}
b_{i}^{*}(x_1,x_2)(I_{1}) = \lambda b_{i}^{*}(x_1,x_2)(I_{2})
\end{equation}
\begin{equation}
c_{i}^{*}(I_{1}) = c_{i}^{*}(I_{2})
\end{equation}
where $b_i^*(x_1,x_2)$ denotes the abscissa of the ground truth bounding box, and $\lambda$ is a constant transformation factor.

The objective of the proposed dual-view object detection task in this paper can be formally expressed as follows:

\begin{equation}
(B, C) = \textbf{M}(I_{1}, B_{I_1}^*, C_{I_1}^*, I_{2}, B_{I_2}^*(x_1,x_2), C_{I_2}^*)
\end{equation}
where $\textbf{M}$ represents the detection model, which is designed to take as input a pair of views, namely $I_{1}$ and $I_{2}$, along with their associated ground-truth bounding boxes $B_{I_1}^*$, $B_{I_2}^*(x_1, x_2)$, and class labels $C_{I_1}^*$, $C_{I_2}^*$. The primary output of this model comprises two components: a set of predicted bounding boxes $B = \{b_1, b_2, \dots, b_n\}$ and their corresponding class labels $C = \{c_1, c_2, \dots, c_n\}$, both of which are associated with the main-view image $I_1$. 
\subsection{Framework} \label{framework}

Figure \ref{fig-method} provides an overview of the proposed framework. The architecture of AENet consists of two distinct pipelines: the main view and the auxiliary view. In the main view pipeline, the training process is conducted independently of the auxiliary view, resembling the traditional object detection model paradigm. Moreover, we employ a specialized approach that involves cropping individual instances from the main view image using the ground-truth boxes. These expert models, which are trained to specialize in detecting ``challenging categories", are stored for subsequent use in the auxiliary view detection pipeline. In the auxiliary view detection pipeline, we utilize saliency detection techniques to generate approximate location predictions for objects in the auxiliary view image. These location predictions, in combination with the category labels obtained from the main view pipeline, are used to train the auxiliary view detector. Then according to the location relation, we crop the essential instance of the main view image. The cropped image patch is then passed through the expert models trained earlier of main view to refine the prediction.

Finally, we integrate the detection results generated by both the main view and auxiliary view pipelines. By combining the outputs from both branches, AENet is able to deliver enhanced detection performance, with improved accuracy and robustness, particularly in handling occlusions, complex backgrounds, and difficult object categories.
\subsection{Main View Detection Pipeline}\label{main-view}

This pipeline focuses on leveraging the traditional object detection paradigm, utilizing the main view image $I_{1}$ for training and detection. The primary objective of this pipeline is to predict bounding boxes $B = \{b_1, b_2, \dots, b_n\}$ and class labels $C = \{c_1, c_2, \dots, c_n\}$ for the objects present in $I_{1}$. The feature extraction backbone $F_{main}$ processes the main view image $I_{1}$ to generate feature maps, $F_{main}(I_{1}) \rightarrow M_{main}$, where $M_{main}$ represents the feature maps extracted from $I_{1}$. The detection heads, denoted by $\mathcal{D}_{main}$, then predict the bounding boxes and class labels:
\begin{equation}
\vspace{-0.05in}
(B, C) = \mathcal{D}_{main}(M_{main})
\end{equation}

The objective loss function $\mathcal{L}_{main}$ to minimize, consists of a classification loss $L_{cls}$ and a regression loss $L_{reg}$:
\vspace{-0.05in}
\begin{equation}
\vspace{-0.05in}
\mathcal{L}_{main} = \sum_{i=1}^{n} L_{cls}(c_i, c_i^*) + \sum_{i=1}^{n} L_{reg}(b_i, b_i^*)
\end{equation}
where $c_i^*$ and $b_i^*$ represent the ground-truth class label and bounding box for the annotated prohibited item object $i$.

To handle ``hard categories", we employ a specialized cropping strategy. During training, individual object instances are cropped from $I_{1}$ using the ground-truth bounding boxes $B_{I_1}^* = \{b_1^*, b_2^*, \dots, b_n^*\}$. The cropped image patches, denoted by $\{I_{1}^{crop}(b_1^*), I_{1}^{crop}(b_2^*), \dots, I_{1}^{crop}(b_n^*)\}$, are passed through expert models $\{E_1, E_2, \dots, E_m\}$, each trained specifically to detect ``challenging categories":
\vspace{-0.02in}
\begin{equation}
\hat{c}_i, \hat{b}_i = E_j(I_{1}^{crop}(b_i^*)), \quad \text{for } i \in [1, n], j \in [1, m]
\end{equation}
The expert models output refined predictions, denoted by $\hat{B} = \{\hat{b}_1, \hat{b}_2, \dots, \hat{b}_n\}$ and $\hat{C} = \{\hat{c}_1, \hat{c}_2, \dots, \hat{c}_n\}$ and the expert models will later be utilized during inference of the auxiliary view pipeline for more accurate detection.

\subsection{Auxiliary View Detection Pipeline}\label{auxiliary-view}
Since the auxiliary view lacks direct ground-truth labels, we adopt an indirect training approach based on saliency detection and the cross-view correspondence.
First, a saliency detection function $\mathcal{S}$ is applied to the auxiliary view image to obtain approximate object locations:
\vspace{-0.05in}
\begin{equation}
B_{aux} = \mathcal{S}(I_{2}) = \{b_{aux_1}, b_{aux_2}, \dots, b_{aux_m}\}
\end{equation}
These bounding boxes are assigned class labels transferred from the main view using a correspondence relationship. Specifically, each auxiliary bounding box $b_{aux_i}$ is matched to its corresponding main view bounding box $b_{main_i}$, and the class labels are assigned accordingly:

\vspace{-0.1in}
\begin{equation}
c_{aux_i} = c_{main_i}, \text{for corresponding } b_{aux_i} \text{ and } b_{main_i}
\end{equation}
The auxiliary detector, denoted by $\mathcal{D}_{aux}$, then learns to predict bounding boxes and labels from the auxiliary view feature maps $M_{aux}$ extracted by a feature extractor $F_{aux}$:
\vspace{-0.05in}
\begin{equation}
M_{aux} = F_{aux}(I_{2}), (\hat{B}_{aux}, \hat{C}_{aux}) = \mathcal{D}_{aux}(M_{aux})
\end{equation}
The training objective for the auxiliary view is to minimize the loss $\mathcal{L}_{aux}$, which consists of a saliency-based regression loss $L_{sal}$ and a classification loss $L_{cls}$:
\vspace{-0.05in}
\begin{equation}
\mathcal{L}_{aux} = \sum_{i=1}^{m} L_{sal}(b_{aux_i}, \hat{b}_{aux_i}) + \sum_{i=1}^{m} L_{cls}(c_{aux_i}, \hat{c}_{aux_i})
\end{equation}
\vspace{-0.05in}
During inference, the auxiliary view bounding boxes may overlap. A merging strategy is employed to combine overlapping boxes, leading to more accurate predictions:
\vspace{-0.05in}
\begin{equation}
B_{aux}^{merged} = \textbf{Merge}(\hat{B}_{aux})
\end{equation}
\vspace{-0.05in}
where the function $\textbf{Merge}(\cdot)$ represents an algorithm (such as non-maximum suppression or a weighted average technique) for merging overlapping prediction bounding boxes.

The merged bounding boxes $B_{aux}^{merged}$ and their class labels $\hat{C}_{aux}$ are then used to crop the corresponding regions from the main-view image $I_{1}$ (due to positional invariance of the imaging devices, a corresponding horizontal coordinate relationship exists between objects in these two perspective images). The cropped instances are passed through the corresponding expert models to refine the detection:

\begin{equation}
I_{1}^{crop}(b_{aux_i}^{merged}) \quad \text{for } b_{aux_i}^{merged} \in B_{aux}^{merged}
\end{equation}
\begin{equation}
(\tilde{c}_i, \tilde{b}_i) = E_j(I_{1}^{crop}(b_{aux_i}^{merged})), \text{for } i, j \in [1, m]
\end{equation}

The final set of bounding boxes $B_{final}$ is obtained by combining the bounding boxes from the main view $(B)$ and the auxiliary view $\{\tilde{b}_1, \tilde{b}_2, \dots, \tilde{b}_m\}$. The final set of class labels $C_{final}$ is obtained by combining the class labels from the main view $(C)$ and the auxiliary view $\{\tilde{c}_1, \tilde{c}_2, \dots, \tilde{c}_m\}$:
  
\begin{equation}
B_{u}, C_{u} = B \cup \{\tilde{b}_1, \tilde{b}_2, \dots, \tilde{b}_m\}, C \cup \{\tilde{c}_1, \tilde{c}_2, \dots, \tilde{c}_m\}
\end{equation}

Thus, the final detection results $(B_{final}, C_{final})$ represent the complete set of bounding boxes and labels by combining the contributions from both pipelines, enhancing detection quality by leveraging complementary information.
\subsection{Overall Procedure}\label{training_progress}

Algorithm \ref{alg:avenet_training} summarizes the whole process. The overall training and inference process of AENet involves training both main and auxiliary view pipelines independently and then combining their results. During main view training, $I_{1}$ is used to predict bounding boxes $B$ and class labels $C$, with hard categories refined through expert models. Auxiliary view training utilizes saliency detection on $I_{2}$ to estimate object locations, which are labeled based on correspondence with the main view. During inference, auxiliary bounding boxes are refined, merged, and used to crop instances from $I_{1}$ for further refinement via expert models. Finally, outputs from both views are unified to generate the final bounding boxes and labels $(B_{u}, C_{u})$, ensuring improved accuracy and robustness by leveraging complementary information from both perspectives.

\renewcommand{\algorithmicrequire}{\textbf{Input:}} 
\renewcommand{\algorithmicensure}{\textbf{Output:}}
\begin{algorithm}[!t]
	\caption{Overall Procedure of AENet}
	\label{alg:avenet_training}
	\KwIn{Main view image $I_{1}$, auxiliary view image $I_{2}$, ground-truth boxes $B_{I_1}^*$, class labels $C_{I_1}^*$.}
	\KwOut{Trained detection model $\textbf{M}$, final bounding boxes $B_{u}$, class labels $C_{u}$.}
	
	\textbf{Main View Training:} \\
	Extract feature maps $M_{main}$ from $I_{1}$; \\
 Predict $(B, C)$ using $\mathcal{D}_{main}$ based on $M_{main}$; \\
 Compute loss $\mathcal{L}_{main}$ based on Eq. (5).\\
	\For{each $b_i^* \in B_{I_1}^*$}
	{
		Crop $I_{1}^{crop}(b_i^*)$;\\ 
  Train expert models $E_j$ for hard categories.
	}
	
	\textbf{Auxiliary View Training:} \\
	Obtain $B_{aux} \leftarrow \mathcal{S}(I_{2})$; \\
 Assign class labels $c_{aux_i} = c_{main_i}$ for related boxes.\\
	Extract $M_{aux} \leftarrow F_{aux}(I_{2})$; \\Predict $(\hat{B}_{aux}, \hat{C}_{aux})$ using $\mathcal{D}_{aux}$; \\
 Compute loss $\mathcal{L}_{aux}$ based on Eq. (10).
	
	\textbf{Inference and Refinement:} \\
	Merge boxes: $B_{aux}^{merged} = \textbf{Merge}(\hat{B}_{aux})$.\\
	\For{each $b_{aux_i}^{merged} \in B_{aux}^{merged}$}
	{
		Crop $I_{1}^{crop}(b_{aux_i}^{merged})$; \\
  Refine $(\tilde{c}_i, \tilde{b}_i) = E_j(I_{1}^{crop}(b_{aux_i}^{merged}))$.
	}
	
	Combine final results: $(B_{u}, C_{u}) = (B \cup \tilde{B}, C \cup \tilde{C})$.\\
	
	Obtain trained model $\mathbf{M}$ and final outputs $(B_{u}, C_{u})$.
	
\end{algorithm}

\begin{table*}[!t]
\setlength{\tabcolsep}{1.8mm}
  \small
  \vspace{-0.3in}
  \centering
  \begin{tabular}{c|c|cccccccc|c|ccc}
    \toprule
    \multirow{2}{*}{\centering Method}& \multirow{2}{*}{\centering \textbf{AP\textsubscript{50}}}& \multicolumn{8}{c|}{Simple Categories (\textbf{Same Main-view Detector})}&\multicolumn{4}{c}{Challenging Categories}\\
    \cmidrule(lr{7pt}){3-10} \cmidrule(lr{7pt}){11-14}&&MP & OL & PC1 & PC2 & LA & GL & TA & BL  & $\text{NL}^{*}$& CO & UM & CG  \\
    \midrule 
    
    F-RCNN~\cite{faster}  & 67.8&95.8&67.0&96.8&97.0&76.0&96.4&93.9&90.4&30.1&12.6&45.6&12.2\\
    \rowcolor[rgb]{0.988,0.914,0.914}\textbf{+ours}&\textbf{69.7}$_{\color[rgb]{0.8,0,0}{\uparrow 1.9\%}}$&95.8&67.0&96.8&97.0&76.0&96.4&93.9&90.4&30.1&\textbf{13.3}$_{\color[rgb]{0.8,0,0}{\uparrow 0.7\%}}$&\textbf{65.7}$_{\color[rgb]{0.8,0,0}{\uparrow 20.1\%}}$&\textbf{14.2}$_{\color[rgb]{0.8,0,0}{\uparrow 2.0\%}}$\\
    \midrule
    C-RCNN~\cite{cascade}  &66.6 &95.6 & 63.7 & 96.6 & 97.3 & 72.7 & 96.0 & 93.0 & 89.4 & 21.9& 14.1  & 42.1 & 16.5 \\
    \rowcolor[rgb]{0.988,0.914,0.914}\textbf{+ours}&\textbf{68.6}$_{\color[rgb]{0.8,0,0}{\uparrow 2.0\%}}$&95.6 & 63.7 & 96.6 & 97.3 & 72.7 & 96.0 & 93.0 & 89.4  & 21.9& \textbf{14.5}$_{\color[rgb]{0.8,0,0}{\uparrow 0.4\%}}$ & \textbf{63.8}$_{\color[rgb]{0.8,0,0}{\uparrow 21.7\%}}$ & \textbf{18.5}$_{\color[rgb]{0.8,0,0}{\uparrow 2.0\%}}$\\
    \midrule
    S-RCNN~\cite{sparse} &  69.6&97.3 & 65.1 & 96.4 & 98.2 & 75.5 & 95.7 & 94.8 & 90.2 & 1.0  & 30.4& 61.0  & 29.2 \\
    \rowcolor[rgb]{0.988,0.914,0.914}\textbf{+ours}&\textbf{70.4}$_{\color[rgb]{0.8,0,0}{\uparrow 0.8\%}}$&97.3 & 65.1 & 96.4 & 98.2 & 75.5 & 95.7 & 94.8 & 90.2  & 1.0 & 30.4& \textbf{67.6}$_{\color[rgb]{0.8,0,0}{\uparrow 5.6\%}}$ & \textbf{31.7}$_{\color[rgb]{0.8,0,0}{\uparrow 2.5\%}}$\\
    \midrule
    RetinaNet~\cite{Retina}  & 62.7&96.5&63.9&96.1&97.8&73.4&95.2&94.3&88.4&2.7&7.0&33.9&3.1\\
    \rowcolor[rgb]{0.988,0.914,0.914}\textbf{+ours}&\textbf{65.1}$_{\color[rgb]{0.8,0,0}{\uparrow 2.4\%}}$&96.5&63.9&96.1&97.8&73.4&95.2&94.3&88.4&2.7&\textbf{9.2}$_{\color[rgb]{0.8,0,0}{\uparrow 2.2\%}}$&\textbf{59.0}$_{\color[rgb]{0.8,0,0}{\uparrow 25.1\%}}$&\textbf{5.1}$_{\color[rgb]{0.8,0,0}{\uparrow 2.0\%}}$\\
    \midrule
    CenterNet~\cite{duan2019centernet} &  72.1&97.4 & 66.0 & 96.0 & 98.2 & 76.0 & 94.9 & 95.3 & 89.8  & 10.0& 39.7 & 62.2 & 39.9\\
    \rowcolor[rgb]{0.988,0.914,0.914}\textbf{+ours}&\textbf{72.4}$_{\color[rgb]{0.8,0,0}{\uparrow 0.3\%}}$&97.4 & 66.0 & 96.0 & 98.2 & 76.0 & 94.9 & 95.3 & 89.8  & 10.0 & 39.7&\textbf{65.4}$_{\color[rgb]{0.8,0,0}{\uparrow 3.2\%}}$&39.9\\
    \midrule
    RepPoints~\cite{reppoints}  & 71.9&97.3&69.8&97.4&98.1&78.3&96.9&96.1&92.0&20.8&31.6&53.6&31.2\\
    \rowcolor[rgb]{0.988,0.914,0.914}\textbf{+ours}&\textbf{73.0}$_{\color[rgb]{0.8,0,0}{\uparrow 1.1\%}}$&97.3&69.8&97.4&98.1&78.3&96.9&96.1&92.0&20.8&31.6&\textbf{65.8}$_{\color[rgb]{0.8,0,0}{\uparrow 12.2\%}}$&\textbf{32.4}$_{\color[rgb]{0.8,0,0}{\uparrow 1.2\%}}$\\
        \midrule
    ATSS~\cite{ATSS} &  72.8&97.0 & 67.7 & 96.9 & 97.9 & 76.6 & 95.7 & 95.0 & 89.4 & 37.7 & 21.9 & 60.0 & 37.4\\
    \rowcolor[rgb]{0.988,0.914,0.914}\textbf{+ours}&\textbf{73.4}$_{\color[rgb]{0.8,0,0}{\uparrow 0.6\%}}$&97.0 & 67.7 & 96.9 & 97.9 & 76.6 & 95.7 & 95.0 & 89.4 & 37.7 & 21.9 &\textbf{67.6}$_{\color[rgb]{0.8,0,0}{\uparrow 7.6\%}}$& 37.4\\
    \bottomrule
  \end{tabular}
  \vspace{-0.1in}
    \caption{Performance of various state-of-the-art detection methods with and without the integration of AENet. The first line reports performance on the main-view data, and \textbf{the second line includes the auxiliary view with AENet integrated}. As illustrated, our model improves the detection performance in the challenging categories while maintaining stable performance on the other simple categories. $^{*}$Non-metallic lighters (NL) represent a unique category, as they almost blend in with the background under X-ray imaging, making them exceptionally difficult to detect. Nevertheless, we have chosen to present these results and look forward to further exploration.}
    \vspace{-0.2in}
    \label{exp-table-3}
\end{table*}

\section{Experiments}\label{sec:exp}
In this section, we present a comprehensive evaluation of the proposed AENet, utilizing the LDXray dataset. We begin by detailing the experimental setup in Section \ref{experimental-setting}, followed by an investigation of the integration of AENet with various baseline detection models in Section \ref{sec_model_integration}. Subsequently, Section \ref{ablation_exp} presents four groups of ablation studies designed to isolate and assess the impact of key components of the system. These studies include: (1) An analysis of the integration of different expert models for detecting cropped images in the main-view pipeline (Section \ref{ablation_exp_expert}). (2) The effect of different approximate location techniques used in the auxiliary-view pipeline (Section \ref{ablation_exp_modules}). (3) The impact of various detection models as auxiliary view detectors in the auxiliary-view pipeline (Section \ref{ablation_auxiliary_detector}). (4) A study of how varying confidence score thresholds influence the results in the auxiliary-view module (Section \ref{ablation_exp_conf}).

\subsection{Experimental Setup}\label{experimental-setting} To evaluate the performance of AENet, we use Average Precision (AP) for each category and mean Average Precision (mAP) for overall performance. The AP for each category is calculated using an Intersection over Union (IoU) threshold of 0.5. The mAP is then computed by taking the mean of these AP values across all categories.

All experiments were conducted using the open-source toolbox \texttt{mmdetection}~\cite{mmdetection}, ensuring a consistent and reproducible environment. For all the baseline detection models, we used the default configuration settings provided by \texttt{mmdetection}, and all models were trained and evaluated on two NVIDIA GeForce RTX 4090 GPUs. The LDXray dataset, used in all our experiments, includes both simple and challenging categories, allowing us to evaluate the robustness and generalizability of the AENet across tasks.

\subsubsection{Integration with Various Detection Models}\label{sec_model_integration}
In this experiment, we evaluate the performance of AENet when integrated with various SOTA detection models. The goal is to assess the generalizability of AENet and its ability to enhance different detection architectures. We conduct experiments using several popular detection models, including Faster R-CNN (F-RCNN)\cite{faster}, Cascade R-CNN (C-RCNN)\cite{cascade}, Sparse R-CNN (S-RCNN)\cite{sparse}, RetinaNet\cite{Retina}, CenterNet~\cite{duan2019centernet}, RepPoints~\cite{reppoints}, and ATSS~\cite{ATSS}.

The results, presented in Table \ref{exp-table-3}, show that the integration of AENet consistently improves the mAP across all detection models, with substantial gains in both simple and challenging categories. For example, when integrated with F-RCNN, AENet boosts the mAP by 1.9\%, with significant improvements in challenging categories, such as UM (+20.1\%) and CG (+2.0\%). Similarly, integration with C-RCNN results in a 2.0\% increase in mAP, with even greater improvements in UM (+21.7\%) and CG (+2.0\%). Even for models like RetinaNet, which show lower performance in challenging categories, AENet results in substantial improvements, particularly in UM (+25.1\%) and CG (+2.0\%).

In addition to the improvements in challenging categories, AENet also enhances the overall performance of region-based models such as Faster R-CNN and Cascade R-CNN, as well as anchor-free models like CenterNet and ATSS. For example, ATSS achieves a 0.6\% improvement in mAP when integrated with AENet, with a 7.6\% gain in the challenging UM category. These results demonstrate that AENet is a highly effective and flexible enhancement that can be applied to a wide variety of object detection models, regardless of their underlying architecture.

\subsection{Ablation Studies}\label{ablation_exp}
In this section, we present several ablation studies to analyze the impact of key components of the AENet framework. These studies evaluate the contribution of different techniques used in the auxiliary view module, the integration with various expert detection models, and the effect of varying confidence score thresholds. The results of these experiments are presented in the following sections.

\vspace{-0.1in}
\subsubsection{Expert Models in Main-view Pipeline}\label{ablation_exp_expert} The results of this experiment are shown in Table \ref{exp-table-4}. The primary goal of this study is to identify the most suitable expert model for detecting objects in cropped images within the main-view pipeline. To accomplish this, we evaluated a range of detection models, including all the base models listed in Table \ref{exp-table-3}, with the aim of determining which one performs best in the context of cropped images. Among the models considered, ATSS emerged as the top performer, achieving a mAP of 51.5\%. This result not only outperforms C-RCNN, which achieved a mAP of 50.7\%, and F-RCNN with a mAP of 48.9\%, but also demonstrates a substantial improvement in accuracy across several challenging categories. For instance, ATSS shows a notable improvement in the CO category (+7.4\%), a modest gain in UM (+0.2\%), and a significant boost in CG (+5.4\%). These improvements are particularly important for real-world applications, where detecting objects in difficult or tail categories is often a critical challenge. Taken together, these findings position ATSS as the effective model for analyzing cropped image patches within the main-view pipeline.

\begin{table}[!t]
\vspace{-0.3in}
\setlength{\tabcolsep}{2mm}
  \label{baseline}
  \fontsize{8.9}{13.8}\selectfont
  \centering
  \small
  \begin{tabular}{c|ccc|ccc}
    \toprule
    \multirow{2}{*}{\centering Method}& \multicolumn{3}{c|}{Overall}  &\multicolumn{3}{c}{Categories}\\
    \cmidrule(lr{7pt}){2-4}\cmidrule(lr{7pt}){5-7}
    &mAP&AP\textsubscript{75}&AP\textsubscript{50}&CO  & UM & CG  \\
    \midrule
    F-RCNN~\cite{faster} &48.9 & 49.4 & 80.3 & 76.8 & 94.6 & 66.1\\

    C-RCNN~\cite{faster} &50.7 & 52.0 & 81.7 & 78.0 & 95.8 & 67.5\\

    S-RCNN~\cite{faster} &42.7 & 42.3 & 76.0 & 72.1 & 93.8 & 58.2 \\

        RetinaNet~\cite{faster} &34.7 & 28.4 & 67.2 & 70.4 & 92.8 & 41.7 \\

        CenterNet~\cite{faster} &45.6 & 44.1 & 78.9 & 69.6 & 94.1 & 63.7\\

        RepPoints~\cite{faster} &31.3 & 30.6 & 55.5 & 64.0 & 94.4 & 16.7 \\
        \rowcolor[rgb]{0.988,0.914,0.914}ATSS~\cite{faster} &\textbf{51.5} & \textbf{55.1} & \textbf{83.9} & \textbf{84.2} & \textbf{94.8} & \textbf{72.9} \\
    \bottomrule
  \end{tabular}
  \vspace{-0.05in}
    \caption{Performance comparison of different expert models trained on the cropped images in the main-view pipeline.}
    \vspace{-0.2in}
    \label{exp-table-4}
\end{table}

\vspace{-0.1in}
\subsubsection{Location Approximation in Auxiliary-view}\label{ablation_exp_modules}
The results of this experiment are presented in Table \ref{exp-table-5}. In this study, we evaluate the impact of different location approximation techniques used in the auxiliary-view module of AENet. Specifically, we compare the performance of several methods, including the ``Saliency" technique for approximating object locations. As shown in the table, the saliency technique, which utilizes a grey map approach, outperforms other methods. By incorporating this technique, the mAP increases from 38.2\% to 39.1\%, with notable improvements in challenging categories such as CO (+2.2\%), UM (+25.1\%), and CG (+2.0\%). These results demonstrate that the Saliency technique provides more accurate spatial information, significantly enhancing the auxiliary-view module's ability to refine object localization. As a result, the overall detection performance is improved, confirming the effectiveness of the saliency detection for better location approximation.
\vspace{-0.1in}
\begin{table}[!h]
\vspace{-0.05in}
\setlength{\tabcolsep}{2mm}
  \label{baseline}
  \small
  \centering
  \begin{tabular}{c|ccc|ccc}
    \toprule
        \multirow{2}{*}{\centering  Threshold}& \multicolumn{3}{c|}{Overall}  &\multicolumn{3}{c}{Categories}\\
    \cmidrule(lr{7pt}){2-4}\cmidrule(lr{7pt}){5-7}
    &mAP&AP\textsubscript{75}&AP\textsubscript{50}&CO  & UM & CG  \\
    \midrule 
     
    Single-view&38.2&41.0&62.7&7.0&33.9&3.1  \\
    Conversion&38.2&41.1&62.7&7.5&33.9&3.1\\
    \rowcolor[rgb]{0.988,0.914,0.914}Saliency (\textbf{ours})&\textbf{39.1}&\textbf{41.4}&\textbf{65.1}&\textbf{9.2}&\textbf{59.0}&\textbf{5.1} \\

    \bottomrule
  \end{tabular}
  \vspace{-0.05in}
    \caption{Performance comparison of detection methods with different approximate location techniques in auxiliary-view module.}
    \vspace{-0.2in}
    \label{exp-table-5}
\end{table}
\subsubsection{Auxiliary View Detectors}\label{ablation_auxiliary_detector} The results are shown in Table \ref{exp-table-6}. In this experiment, we explore how the choice of detection model as the auxiliary view detector impacts performance. We evaluate several models for this task, assessing how their performance in the auxiliary-view pipeline influences the overall results.

The results show that using a specialized detector in the auxiliary view module can improve the mAP performance. The best results were observed when using the C-RCNN detector, which led to a 26.7\% mAP performance, particularly in challenging categories like UM (78.6\%) and CG (29.0\%). These results underscore the importance of selecting an appropriate detector for the auxiliary view, emphasizing the proposed model's ability to detect challenging categories, which leads to a higher recall of items during detection.
\vspace{-0.1in}
\subsubsection{Confidence Score Thresholds}\label{ablation_exp_conf} 
\begin{figure}[!h]
  \centering
  \vspace{-0.05in}
\includegraphics[width=0.9\linewidth]{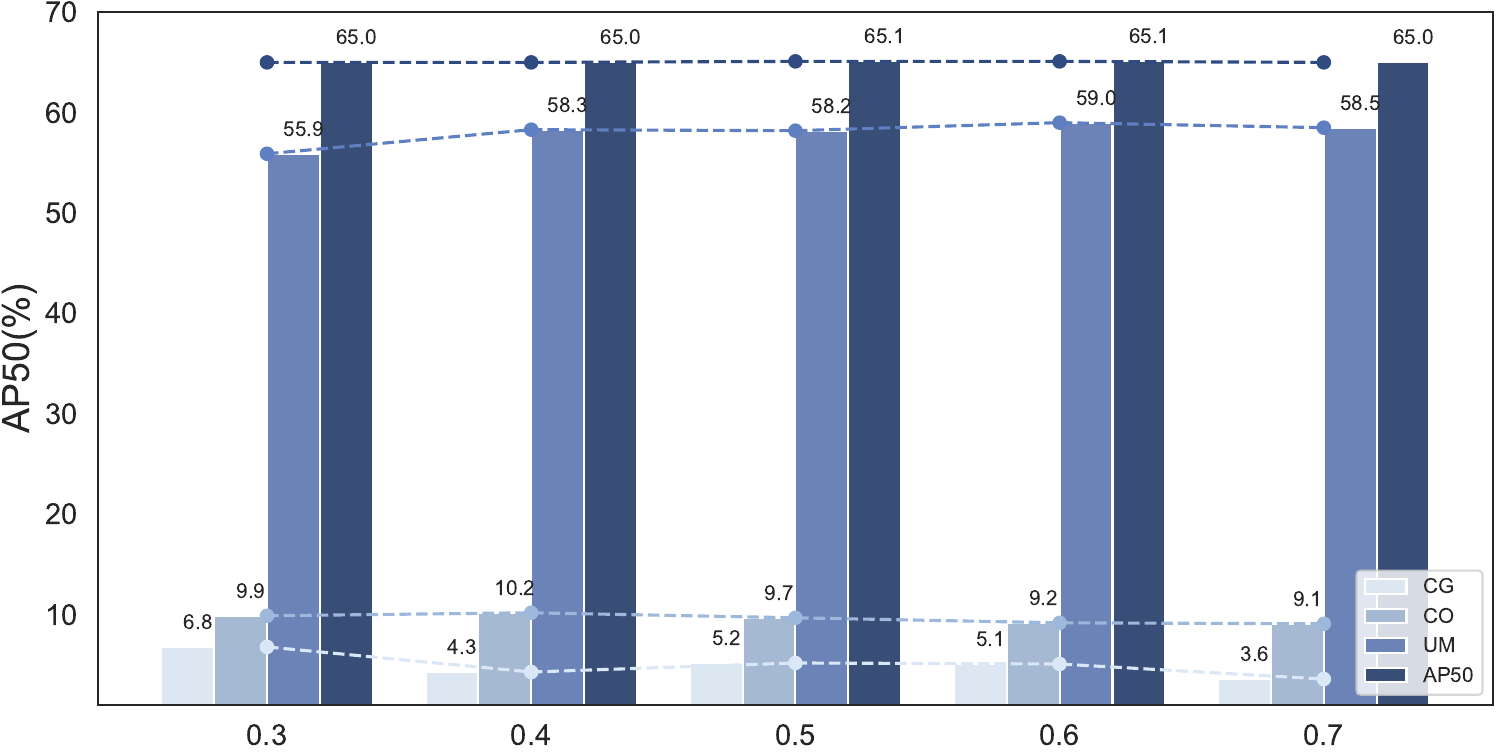}
\vspace{-0.1in}
   \caption{Impact of varying confidence score thresholds on the performance of the auxiliary view module, showing the optimal threshold for improving mAP and category-specific results.}
   \vspace{-0.1in}
   \label{fig:thresholds}
\end{figure}

The results are shown in Figure \ref{fig:thresholds}. In this section, we analyze how varying confidence score thresholds influence the auxiliary-view module’s performance. By adjusting the confidence threshold, we examine how the trade-off between precision and recall impacts the final results.

\begin{table}[!t]
\vspace{-0.3in}
\setlength{\tabcolsep}{2mm}
  \label{baseline}
  \fontsize{8.9}{13.8}\selectfont
  \centering
  \small
  \begin{tabular}{c|ccc|ccc}
    \toprule
    \multirow{2}{*}{\centering Method}& \multicolumn{3}{c|}{Overall}  &\multicolumn{3}{c}{Categories}\\
    \cmidrule(lr{7pt}){2-4}\cmidrule(lr{7pt}){5-7}
    &mAP&AP\textsubscript{75}&AP\textsubscript{50}&CO  & UM & CG  \\
    \midrule 
    F-RCNN~\cite{faster} &24.3 & 20.2 & 49.0 & 9.0 & 75.8 & 22.3 \\

    \rowcolor[rgb]{0.988,0.914,0.914}C-RCNN~\cite{faster} &\textbf{26.7} & \textbf{24.2} & \textbf{53.8} & \textbf{8.7} & \textbf{78.6} & \textbf{29.0}\\

    S-RCNN~\cite{faster} & 8.2 & 7.2 & 16.1 & 4.8 & 25.3 & 6.8\\

        RetinaNet~\cite{faster} &20.1 & 16.7 & 43.4 & 4.5 & 76.3 & 10.4 \\

        CenterNet~\cite{faster} & 22.3 & 19.6 & 47.0 & 9.2 & 78.2 & 15.7 \\

        RepPoints~\cite{faster} &4.2 & 3.1 & 10.1 & 4.5 & 16.7 & 3.5 \\
        ATSS~\cite{faster} &25.6 & 23.2 & 49.7 & 11.5 & 79.3 & 20.1 \\
    \bottomrule
  \end{tabular}
  \vspace{-0.05in}
    \caption{Comparison of performance across different auxiliary view detectorss \textbf{integrated with the proposed AENet}.}
       \vspace{-0.2in}
    \label{exp-table-6}
\end{table}

As shown in Figure \ref{fig:thresholds}, setting the threshold to 0.6 yields the highest mAP of 39.1, with significant improvements in the CO and UM categories. Lower thresholds (\emph{e.g.}, 0.3 or 0.4) result in a decrease in overall performance, likely due to an increase in false positives, while higher thresholds (e.g., 0.7) result in a drop due to missing true positives. These findings highlight the importance of setting an optimal confidence score threshold to maximize the contribution of the auxiliary-view module to the final detection results.
\vspace{-0.05in}
\section{Conclusion}\label{Section:conclusion}
\vspace{-0.05in}
This paper introduces the Large-scale Dual-view X-ray (LDXray) dataset and the Auxiliary-view Enhanced Network (AENet). By leveraging both main and auxiliary views of X-ray images, AENet addresses the limitations of single-view detection, particularly for challenging categories. The extensive experiments conducted on the LDXray dataset demonstrate the effectiveness of the dual-view mechanism, showcasing its ability to enhance detection accuracy. The proposed dataset and framework not only offer a valuable resource for further research but also hold the potential to improve real-world security inspection.

{
    \newpage
    \small
    \bibliographystyle{ieeenat_fullname}
    \bibliography{main}
}


\end{document}